# StrokeSave: A Novel, High-Performance Mobile Application for Stroke Diagnosis using Deep Learning and Computer Vision

**Ankit Gupta**

Thomas Jefferson High School for Science and Technology

Alexandria, Virginia, USA

## Abstract

According to the WHO, Cerebrovascular Stroke (CS) is the second largest cause of death worldwide. Current diagnosis of CS relies on labor and cost intensive neuroimaging techniques, unsuitable for areas with inadequate access to quality medical facilities. Thus, there is a great need for an efficient diagnosis alternative. StrokeSave is a platform for users to self-diagnose for prevalence to stroke. The mobile app is continuously updated with heart rate, blood pressure, and blood oxygen data from sensors on the patient's wrist. Once these measurements reach a threshold for possible stroke, the patient takes facial images and vocal recordings to screen for paralysis attributed to stroke. A custom designed lens attached to a phone's camera then takes retinal images for the deep learning model to classify based on presence of retinopathy and sends a comprehensive diagnosis. The deep learning model, which consists of a RNN trained on 100 voice slurred audio files, a SVM trained on 410 vascular data points, and a CNN trained on 520 retinopathy images, achieved a holistic accuracy of 95.0% when validated on 327 samples. This value exceeds that of clinical examination accuracy, which is around 40%-89%, further demonstrating the vital utility of such a medical device. Through this automated platform, users receive efficient, highly accurate diagnosis without professional medical assistance, revolutionizing medical diagnosis of CS and potentially saving millions of lives.

# Introduction

According to the WHO, cerebrovascular stroke is the second largest cause of death worldwide, accounting for more than 5 million deaths in 2017 alone. Conventional detection methods involve physician administered exams including blood tests, angiograms, carotid ultrasounds, CT, and MRI scans. These neurological scans such as MRI are first of all costly, and are generally obtained 12 hours after the onset of symptoms, thus preventing patients from receiving treatment in the optimal timeframe. Thus, early, automated diagnosis methods using other biomarkers of stroke have been a central focus on cerebrovascular accident (CVA) research. Currently, there are no existing portable stroke screening platform available to the public. This is a dire issue requiring immediate action, since over 140,000 people die of stroke in the US and in every 40 seconds, someone is afflicted with stroke. Since current diagnosis methods yield accuracies up to 80% and stroke diagnosis is largely inaccessible in areas without access to quality medical facilities, a holistic patient centered early diagnosis application is critical to address this pressing issue. Moreover, recent advances in the applications of deep learning and computer vision, backend databases, and mobile application development allow us to engineer a timely, inexpensive, accurate stroke detection platform with far higher accuracy than current medical professionals.

## Background

Cerebrovascular accident (CVA) is a disease in which blood flow to a part of the brain is stopped either by a blockage or rupture of a blood vessel. Using diagnosis techniques such as blood tests, angiograms, carotid ultrasounds, CT scans, and echocardiograms, physicians determine the best course of treatment, which usually involves a combination of clot busters (tPA) and procedures such as thrombectomy.

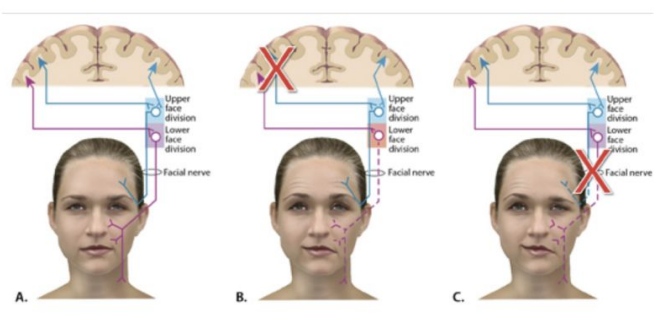

*Figure 1.* Pathway of the facial nerve(A) / effects of injuries to the cortex(B) and the facial nerve(C)

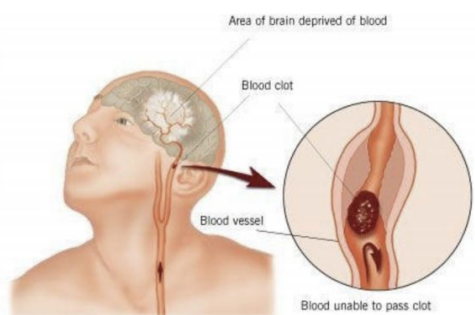

*Figure 2.* Stroke induced vocal paralysis http://fortworthent.net/voice-problems/vocal-cord-disorders/

Transient ischemic attacks (TIA) are temporary blockages in the brain often caused by a buildup of plaque in major arteries. Several symptoms of TIA, such as a sagging facial features, arm weakness, and speech difficulties, can be assessed to determine whether the patient has a stroke, and further steps can be taken to prevent a major stroke in the future. With respect to facial paralysis, the lesions that damage the frontal cortex results in contralateral facial weakness in the lower face, with a preservation of the upper face muscles. Additional factors in prediction of onset of an impending stroke include vascular data manifested in heart rate, blood pressure, and blood oxygen level measurements. Spikes in heart rate and blood pressure are associated with the sudden blockage of a blood vessel, as is a sudden drop in blood oxygen level.

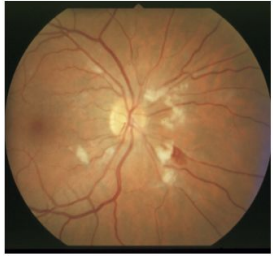 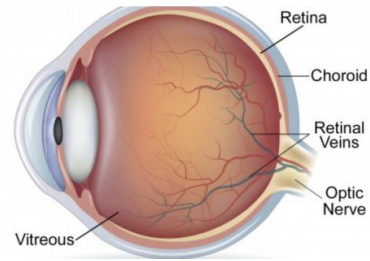

*Figure 3.* Hypertensive retinopathy effects http://www.tedmontgomery.com/the_eye/eyephotos/HypertensiveRetinopathy-3.html

*Figure 4.* Retinal Vein Occlusion Components https://www.vmrinstitute.com/vein-occlusions/

Furthermore, to predict the onset of an impending stroke, researchers at the American Atherosclerosis Society have uncovered several retinal and atrial biomarkers that can be attributed to an impending stroke, including hypertensive retinopathy. Hypertensive retinopathy lesions mainly result from small arteriosclerosis and uncontrolled hypertension, which lead to retinal ischemia and breakdown of blood retinal barrier. As the retina is the window to cerebral conditions, changes in the retina directly mirror primary cerebral changes characteristic of an impending stroke, such as the increased vessel permeability due to blood brain barrier breakdown. As such, assessing presence of retinopathy like symptoms is essential to assess a long term risk for stroke.

Machine Learning (ML) is a subset of Artificial Intelligence (AI) that involves giving automated systems the ability to "learn" data rather than being explicitly programmed.The Neural Network (NN) is a machine learning algorithm modeled after the brain composed of layers of 'neurons connected by 'synapses'. As data is propagated through the NN, weights and biases are propounded, further allowing the Neural Network to learn data given, forming a ML model. Computer Vision (CV) is an interdisciplinary field within the field of computer science that deals with how computers can gain a high-level understanding of digital images or videos.

Deep Learning (DL) is a subset of the field of ML that investigates integrating multiple Machine Learning algorithms together to learn data representations rather than task-specific algorithms. Furthermore, combining Deep Learning-based models, such as CNNs, RNNs, and SVMs, combined with powerful CV algorithms such as the Active Appearance Model (AAM), yield exceptional accuracies.

## StrokeSave Platform Design

In order to address to issue, our platform, StrokeSave, is built upon the fact that most areas have access to a wireless, mobile device, lens attachment, and medical grade device which can measure vascular data such as blood pressure, blood oxygen, and heart rate, all costing less than a combined $100. SAS is basically a robust android diagnosis application that incorporates a 3 tier diagnosis procedure for predicting the risk of the onset of stroke, ensuring that patients can receive treatment in an optimal timeframe. Our platform utilizes real-time vascular patient data along with Deep Learning and Computer Vision to produce an efficient diagnosis from specific machine learning analysis of common stroke symptoms, including high blood pressure, irregular heart rhythms, hypoxia, facial paralysis, vocal paralysis, and hypertensive retinopathy.

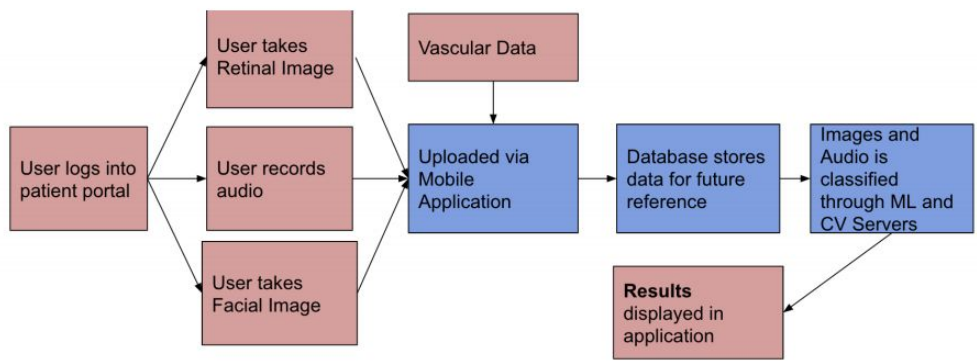

Figure 5. User Interaction with the StrokeSave mobile application platform and Backend Platform Design

These 4 main factors form the basis to the app's diagnostic capabilities. With respect to vascular data, a systolic pressure of around 180 mmHg is indicative of a hypertensive emergency characteristic of stroke. For heart rate, a reading of ~100 beats per minute typically indicates presence of tachycardia that is strongly indicative of stroke. For facial paralysis, stroke induced facial paralysis mostly results in localized facial sagging affecting primarily the contralateral facial muscles. Facial paralysis along with dysarthria(vocal paralysis) are common symptoms that can be seen during the onset of stroke. Since both short term and long term solutions have there own benefits, I seek to implement a holistic, hybrid platform that analyzes both the short and long term risk for stroke.

## Methods

### Deep Learning Architecture

StrokeSave analyzes 4 main factors to assess the risk for stroke:

**Vocal Paralysis Recognition:** The goal of the slurred voice recognition is to address the stroke-induced vocal paralysis inherent in patients at the onset of a stroke. In order to create a deep learning-based model for vocal paralysis, I amassed a dataset of 150 voice slurred audio files from the TORGO Database of Dysarthric Articulation. I split the dataset into training and validation sets of 100 .wav audio files and 50 files respectively.

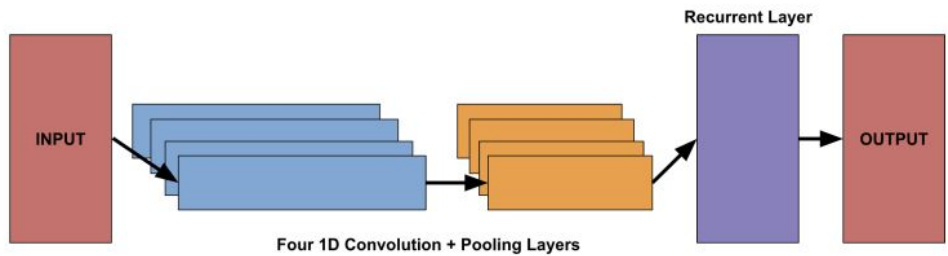

*Figure 6.* Vocal Paralysis recognition CNN and RNN Structure comprised of 4 1D Convolitions and Pooling Layers and a Recurrent Layer.

The .wav files are stored 1D graphical representations, meaning that each unit of a .wav file represents an instant with the measurement of the height of the wave. When the user records their voice using the mobile application and sends it via Firebase, the machine learning model retrieves the audio file as .wav file, runs it through a low pass filter to reduce background noise, then runs it through a hybrid RNN and CNN Structure that are able to learn both spatial and temporal pieces of data.

**Vascular Data Stroke Detection:** The goal of the vascular data-based stroke detection is to address the common, more sudden, vascular effects of Stroke on patients, including heart rate, blood pressure, and blood oxygen level. The data from the user, who wears the blood pressure/oxygen sensors is sent to the app. Using datasets such as those from the NCBI and NIH, I obtained a dataset of 700 vascular data points, which I split into 400 data points for training and the remaining 300 for validation. I designed and ran a Support Vector Machine (SVM) model, using the 400 data points, for which I built a linear regression separating data from patients that had stroke at the time and those that did not.

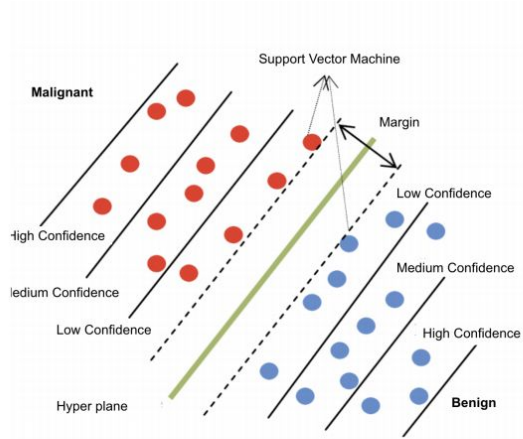

*Figure 7.* Support Vector Machine (SVM) structure implemented for Vascular Data Stroke Detection. Each point represents information from blood oxygene, heart rate, and of blood pressure, while the regression lines separate data points for prediction.

**Hypertensive Retinopathy Detection:** The goal of the hypertensive retinopathy detection is to determine whether the image that the user has submitted of the user's eye using the lens attachment is indeed representative of hypertensive retinopathy, a common symptom present in patients with stroke. Compounding data from the NIH eyeGENE database, I created a dataset of 200 images, 150 of which were used for training and 50 of which were used for validation. After running several pre-processing algorithms on the hypertensive retinopathy detection to get rid of image-based noise, I designed and ran a machine learning-based deep convolutional neural network, which learned, over time, to determine distinguishing features between eye images without hypertensive retinopathy.

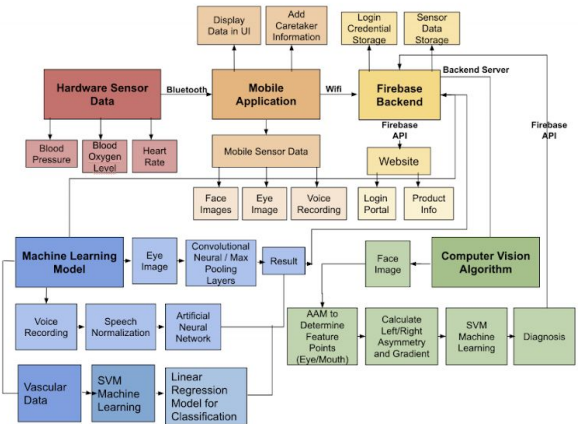 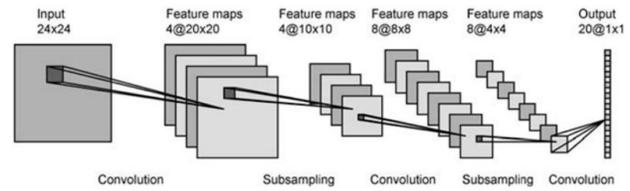

Figure 8. Integrated Stroke Detection pipeline including Deep Learning, CV algorithms, all detection frameworks, database, and front-end sensor hardware and User Interface.

Figure 9. Hypertensive Retinopahy detection Deep Convolutional Neural Network framework implementing specified feature maps of convolutions and subsampling.

**Stroke Detection:** Ultimately, StrokeSave returns an accurate output for percent confidence that the user is at risk for stroke. It uses a SVM implementation that inputs the percent confidence that the user has a slurred voice, vascular data, chance the patient has Hypertensive Retinopathy, and chance the user has facial paralysis, further outputting both whether or not the user is at risk for stroke and the percent risk the user is at during the current time. The algorithm was trained on 300 inputs and validated on 200 inputs.

**Computer Vision Algorithm**

Facial Paralysis Recognition: The goal of the facial paralysis recognition system is to detect the facial paralysis present in patients with stroke. Once a patient takes and submits a picture of their face, the image is sent to Firebase, and the image is retrieved by the Computer VIsion algorithm. In order to compute whether the picture received in the CV architecture was representative of a patient with facial paralysis, I used the Active Appearance Model (AAM) algorithm, a statistical algorithm which matches a statistical model of object and shape to a new image, commonly used in the field of medical image interpretation.

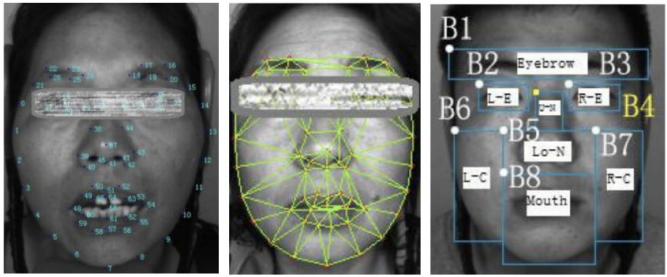

*Figure 10.* Positions of 68 key points on the face, fitting the result using the AAM, and facial regions. https://www.ncbi.nlm.nih.gov/pubmed/25226980

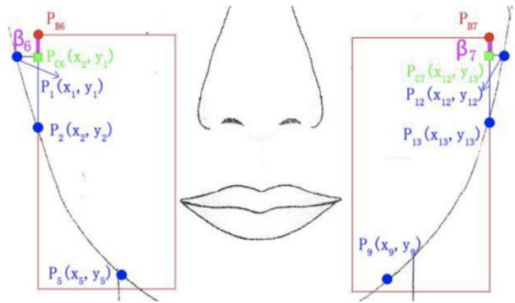

*Figure 11.* Visualization of Dynamic Displacement with respect to the cheek and nose regions, including αn and βn. https://www.ncbi.nlm.nih.gov/pubmed/25226980

Furthermore, I implemented the AAM algorithm by first locating 68 points on the patient's face using the Facial Points Annotation database, which divide the image of the patient's face into eight main regions. Next, the algorithm uses knowledge of the lips and nose being at the center of the face to calculate dynamic displacements from the positions of key points on the organs. The specific dynamic displacements between the cheek regions and the nose are calculated and sent to a Support Vector Machine (SVM) algorithm that determines the difference in dynamic displacement between each of the cheeks and uses a logistic regression to classify whether the dynamic displacements are representative of generalized cases of facial paralysis. Based on the input factors, the SVM model returns an output of whether confidence is higher if the image is of a face with paralysis or a regular face.

# Evaluation

## Results and Discussion

After the deep learning model was trained, the model was tested on a validation data set, and its classification performance was measured using four main statistical measurements for machine learning programs: F-beta score, Precision, Sensitivity, and Accuracy. The definitions are below.

$$F_\beta = \frac{2}{\frac{1}{Precision} + \frac{1}{Sensitivity}} \qquad Sensitivity = \frac{TP}{TP+FP} \qquad Precision = \frac{TP}{TP+FN} \qquad Accuracy = \frac{(TP + TN)}{(TP+FP+TN+TP)}$$

In the above definitions TP means the number of true positives, FP means the number of false positives, and FN is the number of false negatives. The measurements and associated values are displayed in the following table, which accurately depicts measurements and values for each architecture applied.

| Deep Learning/Computer Vision Architecture Applications | | | | | | | |
|---|---|---|---|---|---|---|---|
| Hypertensive Retinopathy (CNN) | | Slurred Speech (RNN/CNN) | | Vascular Data (SVM) | | Facial Paralysis (AAM/SVM) | |
| Measurement | Value | Measurement | Value | Measurement | Value | Measurement | Value |
| Precision | 0.917 | Precision | 0.923 | Precision | 0.951 | Precision | 0.882 |
| Sensitivity | 0.917 | Sensitivity | 0.96 | Sensitivity | 0.869 | Sensitivity | 0.957 |
| F-Beta | 0.917 | F-Beta | 0.941 | F-Beta | 0.908 | F-Beta | 0.918 |
| Accuracy | 92.00% | Accuracy | 94.00% | Accuracy | 90.25% | Accuracy | 92.00% |

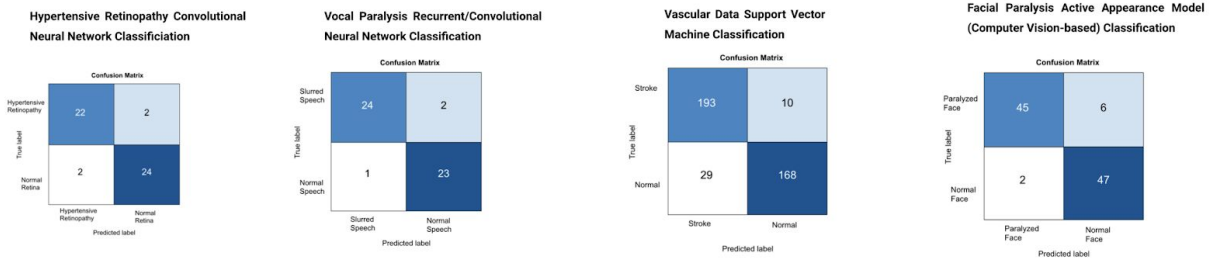

Furthermore, the holistic deep learning architecture that utilizes all four of these architectures with a different dataset has precision, sensitivity, and accuracy values as follow.

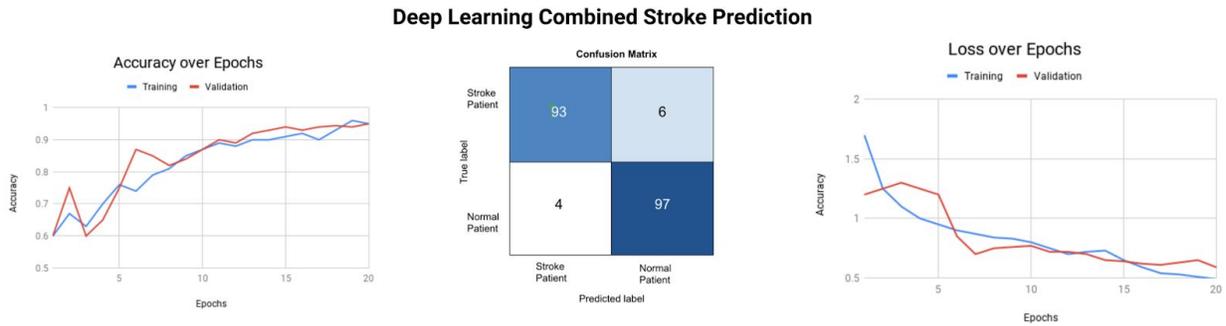

As shown by the evaluation tables, the calculated sensitivity and precision values are consistently similar. An equal number of True Negative and False Positive values combined with high accuracies yields a reliable, high-performing, stroke diagnosis tool.

To robustly determine the medical potential of our platform as a stroke diagnosis tool, I are partnering with the Johns Hopkins School of Medicine as well as Inova Health System to allow patients at risk for stroke to test our diagnosis tool and to receive valuable advice to further improve this highly accurate platform for further evaluation of our holistic stroke prediction model.

## Discussion

Our model's accuracy of 95.0% exceeds that of the best physician's 80% and far exceeds those of beginner to intermediate physicians of 50% to 70%. Our platform is also far more accessible than current neuroimagery techniques, such as CT Scans, MRI Scans, and Cerebral Angiograms, since it is completely noninvasive and neither requires a medical facility nor trained medical professionals. The cost of the entire application is less than $100, compared to expensive medical equipment such as MRI Scans which cost over 20 times as much as the application. Furthermore, our mobile application platform can be used by both caretakers who can follow directions even in a household environment.

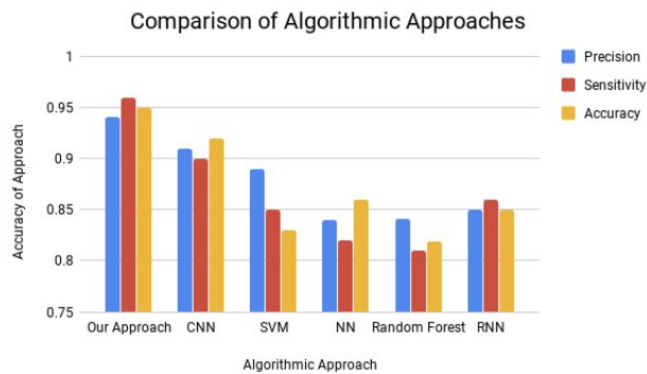

As shown by the chart above, a comparison of the algorithms reveals that the high-level algorithm I implemented is most precise, sensitive, and accurate amongst several other possible algorithmic approaches I could have implemented, further validating the consistency and accuracy of our algorithm as well as reliability and potential in the medical industry.

However, limitations exist. While it is true that our algorithm has high accuracy, one problem I are looking to solve is that of normalizing input from the application such that all images, voice recordings, and data received are representative of the larger dataset that the machine learning models were trained on. Although I did use several computer vision techniques to blur outside backgrounds and further focus on just the face for segmentation, differences in zooming in and shadows can result in highly inaccurate diagnoses. In addition, heavy background noise might be able to pass through the low pass filter and disrupt the slurred voice detection yielding inaccurate results. Thus, to ensure highest accuracy, all recordings should be taken in a quiet room and pictures should be taken in lighter areas with appropriate context.

## **Conclusion**

There are several future endeavors that I wish to pursue in order to further improve this mobile application platform. In terms of structural improvements, it helps improve hardware sensors' connectivity with mobile application for real time analysis and data normalization of the mobile application so it can handle different types of inputs. To further verify the validity of our tool, I would like to actively test the mobile application platform with the help of INOVA/Johns Hopkins Hospital in order to perfect limitations and potential false positives. I would also meet with doctors in the current field to seek advice for improvement of our platform. Currently, I only return whether or not the patient is at risk for stroke and the confidence that a patient has stroke, but as I continue to get more data and the machine learning model becomes more accurate, I would be able to accurately rank symptoms that influence a patient's diagnosis of stroke the most. I are also in the process of retrieving more data from the NIH and other universities to further this research.

      Overall, I developed a high-performance mobile application platform that can diagnose users for Stroke within seconds with 95.0% accuracy while being far more non-invasive, cheaper, time-effective, and accessible than current methods. Such a device truly reveals how a novel and translational engineering project has the potential to help millions of patients around the world receive an accurate diagnosis ahead of time and avoid potential injury or death, further benefiting humanity.